\documentclass[conference]{IEEEtran}
\usepackage{multirow}
\usepackage{balance}
\usepackage{algorithm}
\usepackage{algorithmic}
\usepackage{subfig}

\ifCLASSINFOpdf
   \usepackage[pdftex]{graphicx}
\else
\fi

\begin{document}
%
\title{Ethane: A Heterogeneous Parallel Search Algorithm for Heterogeneous Platforms}

\author{\IEEEauthorblockN{Juli\'an Dom\'inguez}
\IEEEauthorblockA{Departamento de Lenguajes y\\Ciencias de la Computaci\'on\\
Universidad de M\'alaga\\
M\'alaga, Spain\\
Email: julian@lcc.uma.es}
\and
\IEEEauthorblockN{Enrique Alba}
\IEEEauthorblockA{Departamento de Lenguajes y\\Ciencias de la Computaci\'on\\
Universidad de M\'alaga\\
M\'alaga, Spain\\
Email: eat@lcc.uma.es}}

\maketitle

\begin{abstract}

In this paper we present Ethane, a parallel search algorithm specifically designed for its execution on heterogeneous hardware environments. With Ethane we propose an algorithm inspired in the structure of the chemical compound of the same name, implementing a heterogeneous island model based in the structure of its chemical bonds. We also propose a schema for describing a family of parallel heterogeneous metaheuristics inspired by the structure of hydrocarbons in Nature, HydroCM (HydroCarbon inspired Metaheuristics), establishing a resemblance between atoms and computers, and between chemical bonds and communication links. Our goal is to gracefully match computers of different power to algorithms of different behavior (GA and SA in this study), all them collaborating to solve the same problem. The analysis will show that Ethane, though simple, can solve search problems in a faster and more robust way than well-known panmitic and distributed algorithms very popular in the literature.

\end{abstract}

\IEEEpeerreviewmaketitle

\section{Introduction}

Metaheuristics are an important branch of research since they provide a fast an efficient way for solving problems. In many cases, parallelism is necessary, not only to reduce the computation time, but to enhance the quality of the solutions obtained. Many parallel models exist, both for local search methods (LSMs) and evolutionary algorithms (EAs), and even parallel heterogeneous models combining both methods are present in the literature \cite{Alba} \cite{Alba3}.

In a modern lab, it is very common the coexistence of many different hardware architectures. It has been proven that such heterogeneous resources can also be used efficiently to solve optimization problems with standard parallel algorithms \cite{Alba2} \cite{Salto} \cite{Salto2}, but there exist few works about the design of specific parallel models for an heterogeneous environment.

In this paper we propose a parallel search algorithm designed for its execution in a heterogeneous platform. We will also present a draft of a general model for describing a family of heterogeneous metaheuristics specifically designed for its execution in heterogeneous hardware environments, being inspired in the structure of the hydrocarbons that can be found in Nature.

Our contribution is not only methodological, but we also have carried out an analysis in order to study the behavior of our proposal. For our analysis, we have implemented two versions of the algorithm making use of two well-known metaheuristics: steady state Genetic Algorithm (ssGA) and Simulated Annealing(SA). We have compared our proposal with the panmictic versions of these algorithms and with a unidirectional ring of ssGA islands executed on the same hardware infrastructure. Our results show that the running times of our proposal are faster in some cases and more robust in the rest than the reference ssGA ring.

This paper is organized as follows. The next section (Section \ref{section2}) provides a brief review of decentralized and parallel metaheuristics. The Section \ref{section3} explains the proposed algorithm and the model that arises from its chemical inspiration. In Section \ref{section4} we describe some common performance measures which are used in this work. Section \ref{section5} contains the problems, parameters and infrastructure used in our study. The analysis of the tests is exposed in Section \ref{section6}. Eventually, concluding remarks and future research lines are shown in Section \ref{section7}

\section{Decentralized and Heterogeneous Parallel Metaheuristics}\label{section2}

In this section we include a quick review on the existing implementations of decentralized and parallel metaheuristics, as well as on heterogeneity. We also include a description of the metaheuristics used in our heterogeneous algorithm.

Many parallel implementations exist for different groups of metaheuristics. We will focus in two of the more common families of metaheristics: Evolutionary Algorithms (EAs) and Local Search Metaheuristics (LSMs). On the one hand, EAs are population based methods, where a random population is instantiated and enhanced throw a Nature-like evolution process. On the other hand, only one candidate solution is used in LSMs, and it is enhanced by moving through its neighborhood replacing the candidate solution with another one, usually one with a better fitness value. EAs commonly provide a good exploration of the search space, so they are also called exploration-oriented methods. On the contrary, LSMs allow to find a local optima solution and subsequently they are called exploitation-oriented methods. Many different parallel models have been proposed for each method, and here we present the more representative ones.

\subsection{Parallel EA Models}

\begin{figure*}[!t]
\centering
	\includegraphics[width=6in]{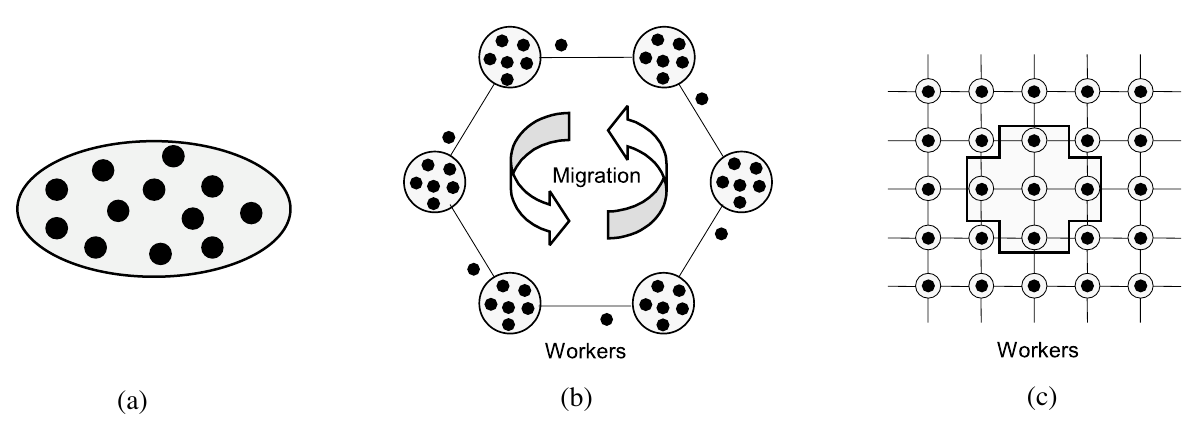}
\caption{A panmictic EA (a), and two structured EAs: distributed (b) and cellular (c).}
\label{pga_models}
\end{figure*}

A panmictic EA applies its stochastic operators over a single population, which makes them easily parallelizable. A first strategy for its parallelization is the use of a master-slave approach where evaluations are performed in parallel but the population, unless divided, is treated as a whole, maintaining its panmictic behavior. It could be interesting for many tasks, but it does not offer the benefits of a structured population. Therefore, we are going to focus in structured populations, which leads to a distinction: we can distinguish between cellular and distributed EAs (Figure \ref{pga_models}).

\begin{itemize}
	\item Distributed EAs (dEA): In the case of distributed EAs, the population is divided into a number of islands that run an isolated instance of the EA (Figure \ref{pga_models}b). Although there is not a single population the sub-populations are not completely isolated: some individuals are sent from one population to another following a migration scheme. With this model there only exists a few sub-algorithms and they are loosely coupled.
	\item Cellular EAs (cEA): In the cellular model, there exists only one population which is structured into neighborhoods, so that an individual can only interact with the individuals inside its neighborhood (Figure \ref{pga_models}c). Different neighborhood structures can lead to a different behavior. With the cellular model there exists a large number of sub-algorithms and they are tightly coupled.
\end{itemize}

\subsection{Parallel LSM Models}

\begin{figure*}[!t]
\centering
	\includegraphics[width=7.3in]{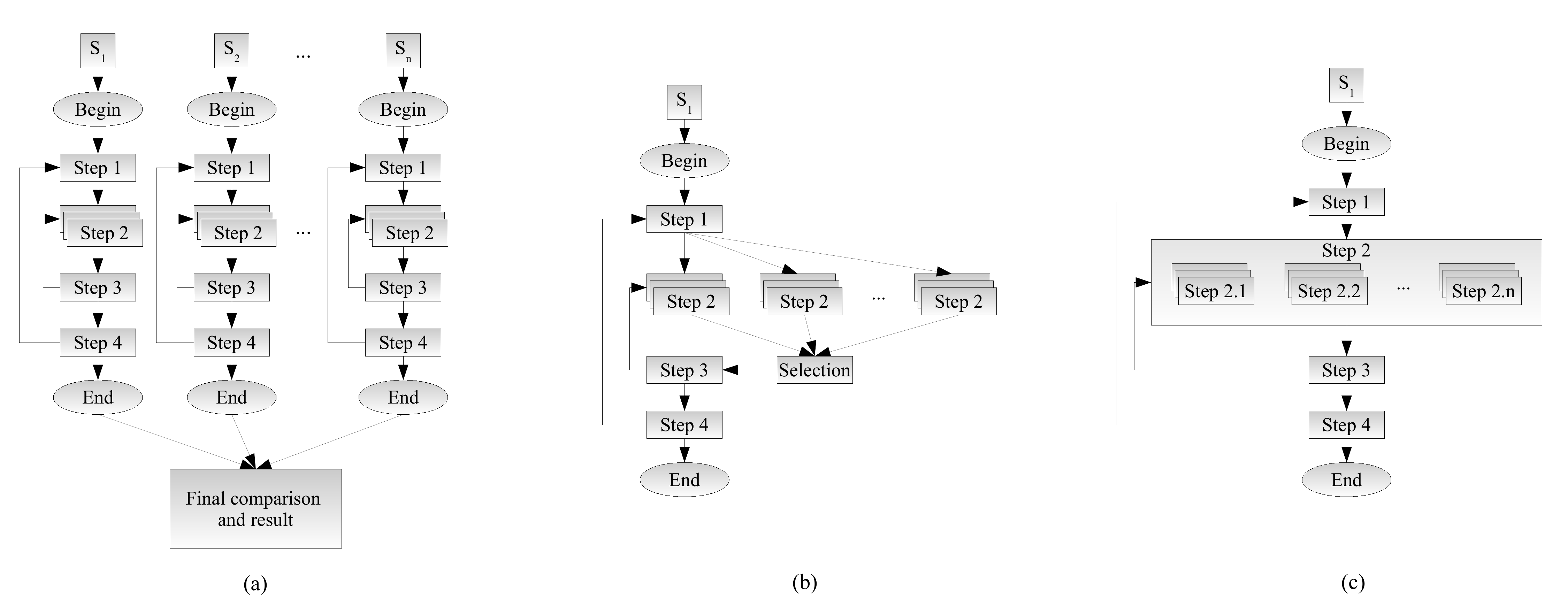}
\caption{Parallel multistart model (a), parallel moves model (b), and move acceleration model (c).}
\label{lsm_models}
\end{figure*}

Many different parallel models have been proposed for LSMs, but there exist three models that are widely extended in the literature: parallel multistart model, parallel moves model, and move acceleration model (Figure \ref{lsm_models}).

\begin{itemize}
	\item Parallel multistart model: In this model, several independent instances of the LSM are launched simultaneously (Figure \ref{lsm_models}a). They can exchange individuals following a migration scheme. This model can usually compute better and more robust solutions than the panmictic version.
	\item Parallel moves model: This model is a kind of master-slave model where the master runs a sequential LSM but, at the beginning of each iteration, the current solution is distributed among all the slaves (Figure \ref{lsm_models}b). The slaves perform a move and return the candidate solution to the master, which selects one of them. This model does not alter the behavior of the algorithm.
	\item Move acceleration model: The quality of each candidate solution is evaluated in a parallel centralized way (Figure \ref{lsm_models}c). It is useful when the evaluation function can be itself parallelized. The move acceleration model does not alter the behavior of the algorithm.
\end{itemize}

In both EAs and LSMs parallel models, each sub-algorithm includes a phase for communication with a neighborhood located on some topology. This communication can be carried out in a synchronous or asynchronous manner. Many works have found advantages in using an asynchronous execution model \cite{Alba4}. Additionally, asynchronism is essential in our study because of the heterogeneous hardware, which could produce bottlenecks, so our model communications are carried out in an asynchronous way.

\subsection{Achieving the Heterogeneity}

In the models presented above, all the sub-algorithms share the same search features. But we can modify the behavior of a parallel metaheuristic by changing the search features between sub-algorithms, obtaining what we call a parallel heterogeneous metaheuristic. Also the hardware being used to run the algorithm can be homogeneous or heterogeneous but we have not to be confused between the hardware platform heterogeneity and the heterogeneous software model. Parallel heterogeneous metaheuristics can be classified in four levels depending on the source of heterogeneity \cite{Alba0}:

\begin{itemize}
	\item Parameter level: At this level, the same algorithm is used in each node, but the configuration parameters are different in one or more of them.
	\item Operator level: At operator level, heterogeneity is achieved by using different mechanisms for exploring the search space, such as different operators.
	\item Solution level: Heterogeneity is obtained using a different encoding for the solutions in each component.
	\item Algorithm level: At this level, each component can run a different algorithm. This level is the most widely used.
\end{itemize}

In this paper we propose an algorithm level parallel heterogeneous metaheuristic which is based in two different methods. We have chosen one method of each of the presented families, LSMs and EAs, in order to obtain a good balance between exploitation and exploration. The used methods are a Genetic Algorithm (GA) and a Simulated Annealing (SA).

GAs are one of the more popular EAs present in the literature. In Algorithm \ref{ga_algorithm} we can see an outline of a panmictic GA. A GA starts with randomly generating a initial population $P(0)$, with each individual encoding a candidate solution for the problem and its associated fitness value. At each iteration, a new population $P'''(t)$ is generated using simple stochastic operators, leading the population towards regions with better fitness values.

\begin{algorithm}\caption{Panmictic Genetic Algorithm}\label{ga_algorithm}
\begin{algorithmic}
	\STATE Generate($P(0)$);
	\STATE Evaluate($P(0)$);
	\STATE t := 0;
	\WHILE {\NOT stop\_condition($P(t)$)} 
		\STATE $P'(t)$ := Selection($P(t)$);
		\STATE $P''(t)$ := Recombination($P'(t)$);
		\STATE $P''(t)$ := Mutation($P''(t)$);
		\STATE Evaluate($P'''(t)$);
		\STATE $P(t+1)$ := Replace($P(t)$,$P'''(t)$);
		\STATE t := t+1;
	\ENDWHILE	
\end{algorithmic}
\end{algorithm}

In our algorithm, we have actually used a special variant of the generic GA called steady state Genetic Algorithm (ssGA) \cite{Syswerda}. The difference between a common generational GA and a ssGA is the replace policy: while in a generational GA a full new population replaces de old one, in a ssGA only a few individuals, usually one or two, are generated at each iteration and merged with the existing population.

Because of its ease of use SA has become one of the most popular LSMs. SA is an stochastic algorithm which explores the search space using a hill-climbing process. A panmictic SA is outlined in Algorithm \ref{sa_algorithm}. SA starts with a randomly generated solution $S$. At each step, a new candidate solution $S'$ is generated. If the fitness value of $S'$ is better than the old value, $S'$ is accepted and replaces $S$. As the temperature $T_k$ decreases, the probability of accepting a lower quality solution $S'$ decays exponentially towards zero according to the Boltzmann probability distribution. The temperature is progressively decreased following an annealing schedule.

\begin{algorithm}\caption{Panmictic Simulated Annealing}\label{sa_algorithm}
\begin{algorithmic}
	\STATE Generate($S$);
	\STATE Evaluate($S$);
	\STATE Initialize($T_0$);
	\STATE k := 0;
	\WHILE {\NOT stop\_condition($S$)} 
		\STATE $S'$ := Generate($S$,$T_k$);
		\IF {Accept($S$,$S'$,$T_k$)} \STATE $S$ := $S'$; \ENDIF
		\STATE $T_{k+1}$:= Update($T_k$);
		\STATE k := k+1;
	\ENDWHILE	
\end{algorithmic}
\end{algorithm}

With the basis of the classic SA, many different versions have been implemented by using a different annealing schedule. In our algorithm we have used the New Simulated Annealing (NSA) \cite{Yao}, which uses a very fast annealing schedule. 

\section{Description of the Model}\label{section3}

In this section we present the particularities of Ethane, as well as we briefly outline the proposal of a generic model being inspired in the chemical compounds called hydrocarbons.

\subsection{Ethane}

With Ethane we propose a Nature inspired heterogeneous parallel search algorithm specifically designed for it execution in a heterogeneous hardware platform. 

Usually, using a generic parallel model within a heterogeneous platform leads to bottlenecks caused by islands with limited resources which can not provide good enough solutions to islands whose populations are much more evolved. Furthermore, many communication topologies does not take care of the underlying hardware architecture and could worsen this problem by overloading these slow islands with too much communication. With Ethane, we propose a communication schema where the most of the communication load is distributed over the fastest nodes of the platform, and the slowest ones are placed as its \emph{slaves}.

\begin{figure}[!t]
\centering
	\includegraphics[width=3in]{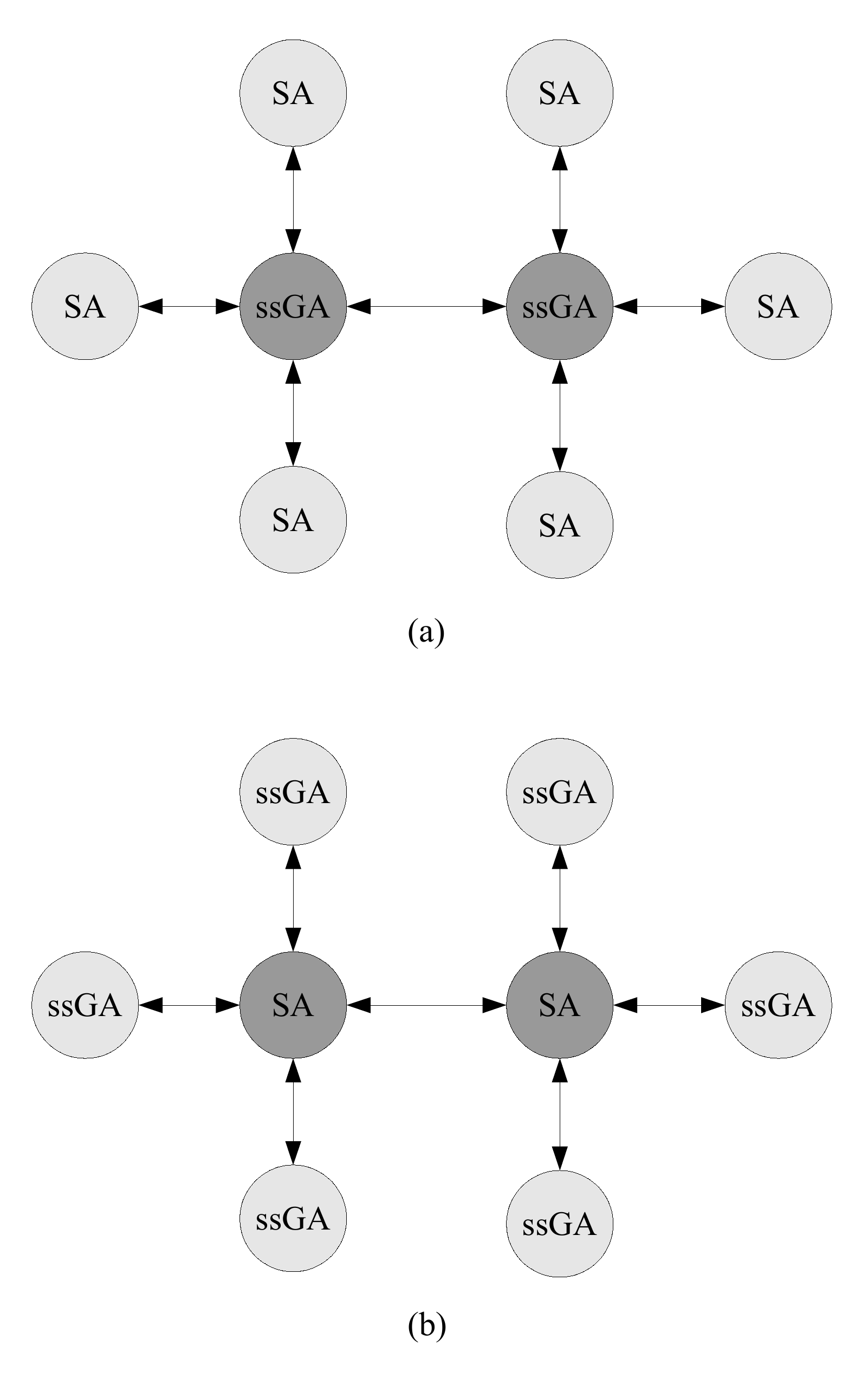}
\caption{Communication schema for Ethane G (a) and S (b)}
\label{hydrocm}
\end{figure}

The chemical compound called ethane consist of two carbon atoms and six hydrogen atoms, joined together with single chemical bonds. We have stablished a resemblance between the atoms and the computers of the platform, and between the chemical bonds and communication channels. In ethane, each carbon atom is bonded to three hydrogen atoms, and there is another bond between both carbon atoms. In our Ethane algorithm, we propose the same schema, using two basic algorithms resembling different atoms, and migration channels resembling bonds. 

For our study, we have implemented two different versions of our algorithm. In Figure \ref{hydrocm} it is shown the schema for the two instances of Ethane studied in this paper. Ethane G (Figure \ref{hydrocm}a) assigns a ssGA sub-algorithm to the central nodes, and a SA sub-algorithm to the \emph{slave} nodes. On the contrary, Ethane S (Figure \ref{hydrocm}b) allocates a SA sub-algorithm in each one of the central nodes, and a ssGA sub-algorithm in the \emph{slave} nodes. With this schema, the most of the communication load falls on the \emph{master} nodes, which are provided with the best hardware, moving some of the load out of the slowest nodes.

\subsection{An Overview of HydroCM}

From this chemical inspiration it arises a generic model based on the different structures of hydrocarbons. We have called it HydroCM (HydroCarbon inspired Metaheuristics). We shaped HydroCM as a generic model for a complete family of parallel heterogeneous metaheuristics. The goal of the model is to provide a schema for the islands and communications of the parallel algorithm to efficiently perform a search over a heterogeneous hardware architecture.

\begin{figure}[!t]
\centering
	\includegraphics[width=3.5in]{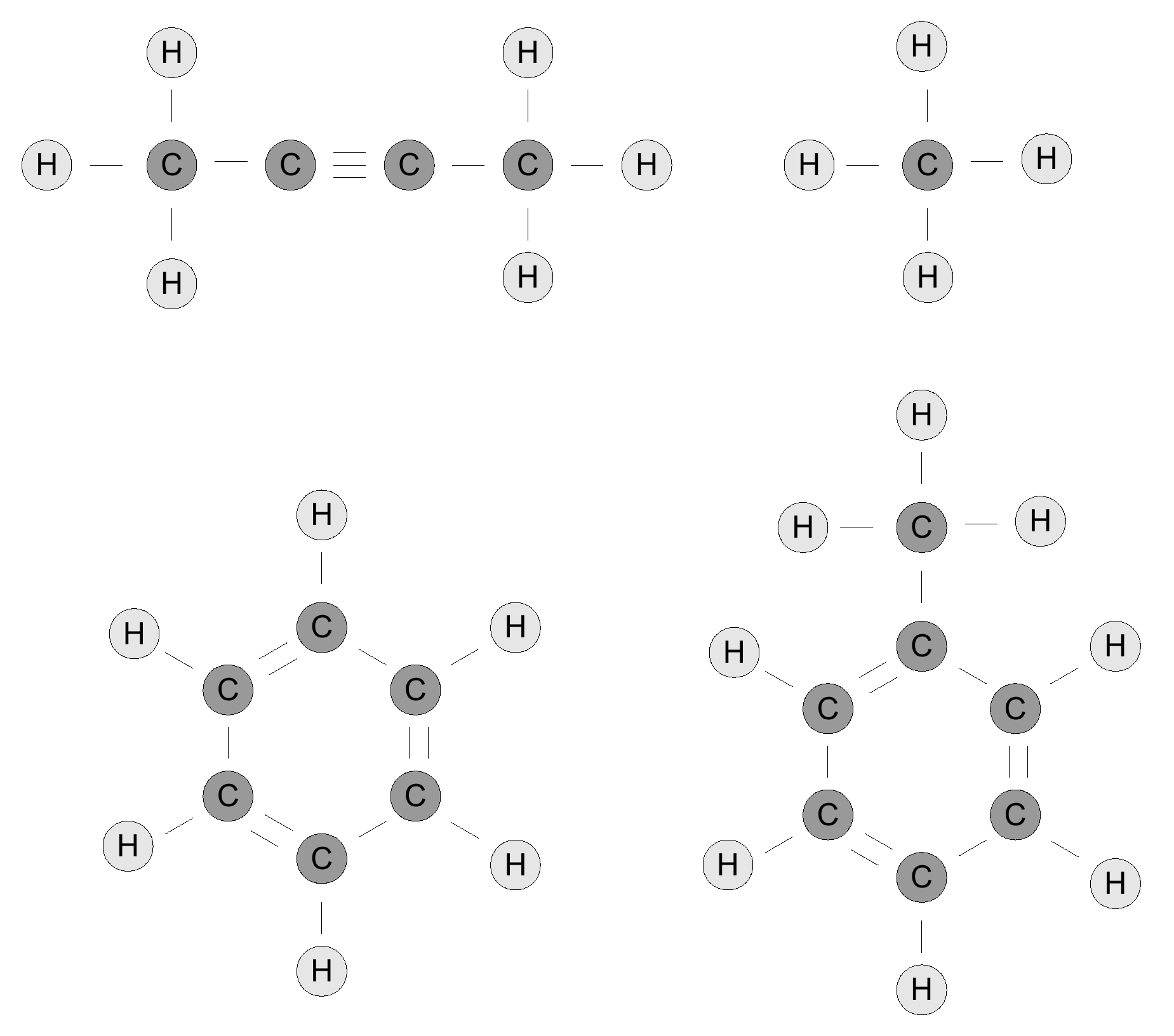}
\caption{Different hydrocarbon configurations that can be found in Nature; their structures are the basis of HydroCM}
\label{hydrocarbons}
\end{figure}

Figure \ref{hydrocarbons} represents some different structures for hydrocarbons as we can find them in Nature. Hydrocarbons are based in only two different atoms, carbon and hydrogen, and each of them can keep a number of bounds, being one for hydrogen and four for carbon.

In our model, we establish a resemblance between computers and atoms in the hydrocarbon. The bonds between atoms have a correspondence to communication channels, and double or triple bonds can be modeled as the amount of information being migrated or, in the case of non-population based algorithms, a higher migration rate. In our model, the fastest machines are associated with central carbon atoms (because of the higher computational effort caused by the migrations) and the slowest ones are associated with hydrogen atoms.

This model provides us with plenty of different schemes for designing a parallel heterogeneous algorithm because of the amount of hydrocarbons present in Nature and their different architectures: linear, ring, branches... obtaining a huge amount of different combinations depending on the number of fast and slow available computers and the topology of the network.

Ethane can be viewed as an instance of HydroCM for an environment composed of eight nodes, where two of them are more powerful than the rest, and making use of ssGA and SA. As well as Ethane is such an instance, we could instantiate many different algorithms depending on the underlying hardware architecture following the model proposed by HydroCM.

\section{Performance Measures and Speedup}\label{section4}

In this section we present the performance measures that are going to be used for assessing the performance of the studied algorithms. The measures that are going to be used are the numerical effort, the total run time and the speedup.

A widely accepted way of measuring the performance of a parallel metaheuristic is to check the number of evaluations of the fitness function needed to locate the optimum. This performance measure is called numerical effort. Numerical effort is widely used in the field of metaheuristics because it removes the effects of the implementation and the platform, but it could be misleading in many cases for parallel methods. Furthermore, the goal of the parallelism is not the reduction of the number of evaluations but the reduction of the running time.

The most significative performance measure for a parallel algorithm is the total run time needed to locate a solution. In a non-parallel algorithm, the use of the \emph{CPU time} is a common performance measure. While parallelizing an algorithm should definitely include any overhead, for example for communication, we are not able to use the \emph{CPU time} as a performance measure. Since the goal of parallelism is to reduce the real time needed to solve the problem, for parallel algorithms it becomes necessary to measure the real run time (wall-clock time) necessary to find a solution.

Because of the non-deterministic behavior of metaheuristics, it is needed to use average values for time and numerical effort. Although 30 runs could provide us a good estimation, we have executed the tests 100 times in order to obtain the statistical analysis.

In our analysis we will also study the speedup. The speedup represents the ratio between sequential and parallel average execution times ($E[T_1]$ and $E[T_m]$ respectively). 

\begin{equation}
s_m=\frac{E[T_1]}{E[T_m]}	
\end{equation}

For the speedup to be a meaningful metric, we have to take care of many aspects for its analysis. Because of the aforementioned non-deterministic behavior of metaheuristics it is necessary to use average times, being these times the wall-clock times. The algorithms run in the single and multiprocessor platform must be exactly the same, thus panmictic algorithms can not be used for the analysis. The algorithms have to be executed until they found the solution or a solution of the same quality. Since in our study we are working over a heterogeneous platform, our reference point is the execution time of the program on the fastest single processor.

\section{Problems, Parameters, and Platform}\label{section5}

In this section we include the basic information necessary to reproduce the experiments that have been carried out for this paper. First we will present the set of benchmark problems used for assessing the performance of our proposal. Second we will briefly explain the parameters used within the sub-algorithms and the underlying hardware and software platform.

\subsection{Benchmark Problems}

In order to assess the performance of our algorithms, we have used two problems in the analysis: the Subset Sum Problem (SSP) \cite{Jelasity} and the Massively Multimodal Deceptive Problem (MMDP) with 6 bits \cite{Goldberg}. 

The SSP problem consists in finding a subset of values $V \subseteq W$ from a set of integers $W= \{ w_1,w_2,...,w_n \} $, such that the subset sum approaches a constant $C$ without exceeding it. We have chosen an instance with 2048 random integer numbers in the range $[0..10^4]$ following a Gaussian distribution. 

MMDP is one of so called deceptive problems. Deceptive problems are specifically designed to make the algorithm converge to wrong regions of the search space, decorrelating the relationship between the fitness of a string and its genotype. In MMDP a binary string encodes $k$ 6-bit sub-problems which contribute with a partial fitness depending on its number of 1's (unitation) following Table \ref{unitation}. We have used an instance with strings of 150 bits so that the global optimum is $k=25$.

\begin{table}[!t]\caption{Bipolar deception (6 bits) sub-function value}\label{unitation}
\begin{center}
\begin{tabular}{ c | c }
	\hline
	\textbf{\#ONES} & \textbf{sub-function value} \\
	\hline
	0 & 1.000000 \\
	1 & 0.000000 \\
	2 & 0.360384 \\
	3 & 0.640576 \\
	4 & 0.360384 \\
	5 & 0.000000 \\
	6 & 1.000000 \\ \hline
	\multicolumn{2}{ }\\
\end{tabular}
\includegraphics[width=3in]{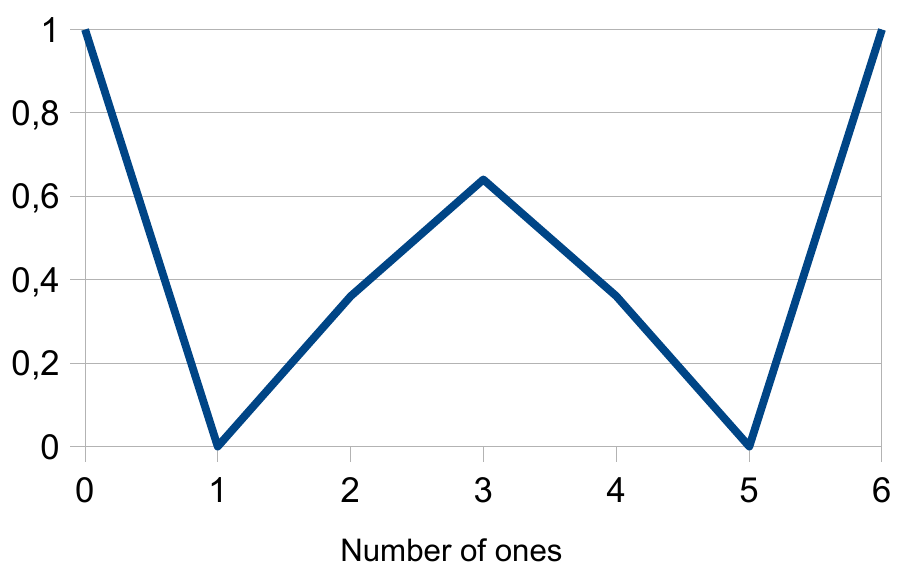}
\end{center}
\end{table}

\subsection{Parameters of the Algorithms and Platform}

The parameters used in every ssGA sub-population are a size of 64 individuals, a crossover probability of 0.8 and a mutation probability of 4.0 divided by the chromosome length. For the SA, we used the same mutation probability. For the SSP the chromosome length is 2048 an in the case of MMDP its length is 150 for both algorithms. In the case of the panmictic ssGA, the population size has been set to 64 individuals because larger populations have performed much worse than smaller ones for the proposed problems, and they have been not able to find the solution of the benchmark problems in a reasonable time.

We have chosen a migration frequency of 50 iterations for all the configurations after several initial preliminary experiments. The number of individuals migrated are 1 in all cases. For the ssGA, the emigrant is randomly selected and the immigrant replaces the worst individual of the population. In the SA, the immigrant is treated as a new move.

The hardware infrastructure used in our analysis consists of 8 different machines: 2 of them have an Intel Core 2 Quad Q9400 @ 2.66GHz processor and 4GB of RAM (namely Type A, fast), the other 6 computers have an Intel Pentium 4 @ 2.4GHz processor and 1GB of RAM (namely Type B, slow). All the computers are managed by a GNU/Linux distribution, being Debian 5.0 for Type A, and SuSE 8.1, Debian 3.1 and Ubuntu 6.10 for Type B. The computers are connected by a Gigabit Ethernet Network. The algorithms have been implemented in Java in order to support both hardware and software heterogeneity. For the purpose of the analysis the version 1.6.0\_01 of the Java Virtual Machine (JVM) is used in all the nodes.

\section{Tests and Analysis}\label{section6}

In this section we analyze the behavior of Ethane, and compare it with the well-known ssGA unidirectional ring. We have analyzed the aforementioned performance measures, being numerical effort, total run time and speedup, as well as the evolution of the fitness.

We have implemented two different algorithms based on Ethane. For the first one, Ethane G, we have provided the Type A computers with a central ssGA island, and Type B computers with a SA island. For the second algorithm, Ethane S, the fastest machines run central SA islands and the slowest ones run ssGA. As we mentioned above, the migration scheme resembles a molecule of ethane as represented in Figure \ref{hydrocm}. In the parallel ssGA used as reference, the islands have been distributed over a unidirectional ring, placing the most powerful computers in the first and fourth place in a sort of MaxSumSort \cite{Branke}. As we do not know the statistical distribution of the data, they have been statistically compared with Mann-Whitney U test.

\subsection{Numerical Effort}

\begin{table}[!t]\caption{Number of evaluations for the tested models and panmictic algorithms}\label{table:evaluations}
\begin{center}
\begin{tabular}{ l | r r | r r}
	\hline
	\multirow{2}{*}{Algorithm} & \multicolumn{2}{c|}{Subset Sum} & \multicolumn{2}{c}{MMDP6} \\
	&Average & Std. Deviation & Average & Std. Deviation\\
	\hline
	Ethane G & \textbf{146418} & \textbf{174433} & 1572735 & 919691 \\
	Ethane S & 202815 & 198696 & \textbf{708231} & \textbf{430353}\\ \hline
	ssGA Ring & 214824 & 239125 & 786583 & 805837\\
	Panm. ssGA & 179792 & 175177 & * & *\\
	Panm. SA & 81737 & 93627 & * & *\\ \hline
\end{tabular}
\end{center}
\end{table}

In Table \ref{table:evaluations} it is represented the numerical effort needed to find the optimum for each algorithm. It can be seen that our proposals performed better than the panmictic algorithms for both problems (in the case of MMPD, panmictic algorithms where not even able to find the optimum in a reasonable time). For the SSP, both Ethane versions performed numerically better than the reference ssGA ring, and one of the instances (Ethane S) performed better even for the MMDP. 

From the point of view of numerical effort, all the differences are statistically significant according to the Mann-Whitney U test. Note that also the standard deviation is better in our algorithms, so that its behavior is more robust. We can see how the panmictic SA has reached the solution with less numerical effort because SA is a fast converging trajectory method, but as we will see with the analysis of the run time, the time needed to find a solution is worse than for the studied parallel models. Since the objective of our model is the reduction of the total execution time let us begin with the analysis of a more meaningful performance metric, the total run time.

\subsection{Total Run Time}

\begin{table}[!t]\caption{Time - ms - for the tested models and panmictic algorithms}\label{table:results}
\begin{center}
\begin{tabular}{ l | r r | r r}
	\hline
	\multirow{2}{*}{Algorithm} & \multicolumn{2}{c|}{Subset Sum} & \multicolumn{2}{c}{MMDP6} \\
	&Average & Std. Deviation & Average & Std. Deviation\\
	\hline
	Ethane G & \textbf{5318} & \textbf{6226} & 9195 & 4942 \\
	Ethane S & 7155 & 6922 & \textbf{3052} & \textbf{1546}\\ \hline
	ssGA Ring & 7453 & 8107 & 3194 & 3380\\
	Panm. ssGA & 30008 & 29387 & * & *\\
	Panm. SA & 13300 & 15443 & * & *\\ \hline
\end{tabular}
\end{center}
\end{table}

Table \ref{table:results} shows the average execution time of each algorithm for each problem until global optimum is reached. As we can see, our proposals performed clearly better than the panmictic algorithms for both problems (remember that the panmictic algorithms where not able to find the optimum for the MMDP) as well as better than the ssGA ring does. 

As we can see in Table \ref{table:results}, Ethane G was the best performing algorithm for the SSP problem. The Mann-Whitney U test gives a $p$-value of 0.0412 for the Ethane G compared to the ssGA ring, so the difference is statistically significant. The average time needed for Ethane G to find a solution is more than 30\% better than for ssGA ring. 

Ethane S was the best algorithm solving the MMDP problem, with an average time slightly better than the ssGA ring, but with a much lower standard deviation, as Mann-Whitney U test confirms with a $p$-value of 0.007. The standard deviation of ssGA is more than twice the standard deviation of Ethane S. This means that the two representative instances of the Ethane family evaluated in this work can be more efficient and more robust/stable than standard sequential and distributed popular algorithms.

\subsection{Speedup}

\begin{table}[!t]\caption{Time - ms - for the tested models in a single processor and its speedup}\label{table:speedup}
\begin{center}
\begin{tabular}{ l | r r | r r}
	\hline
	\multirow{2}{*}{Algorithm} & \multicolumn{2}{c|}{Subset Sum} & \multicolumn{2}{c}{MMDP6} \\
	&Avg. time& Speedup & Avg. time & Speedup\\
	\hline
	Ethane G & \textbf{15995} & \textbf{3.00$\times$} & 41943 & 4.56$\times$ \\
	Ethane S & 17817 & 2.49$\times$ & \textbf{20627} & \textbf{6.76$\times$}\\ \hline
	ssGA Ring & 18137 & 2.43$\times$ & 21227 & 6.64$\times$\\
	\hline
\end{tabular}
\end{center}
\end{table}

In the Table \ref{table:speedup}, we can see a summary of the execution time of the studied algorithms within a single processor and its speedup with respect to the execution in the eight processors heterogeneous platform. As we can see, both versions of Ethane have obtained a better speedup than the ssGA for the SSP, but only Ethane S has achieved a better speedup for the MMDP.

As it is shown in Table \ref{table:speedup}, Ethane G has performed better than the reference ssGA ring even in a single processor in the case of SSP. Even when its performance over a single processor is still better, its speedup is the best of the three models, however, the value for the speedup is not good for any of the algorithms for this problem, being the value for Ethane G 3$\times$. In the case of the MMDP Ethane G has not achieved as good speedup as the ssGA ring.

Ethane S still performed slightly better than the ssGA ring for a single processor for both problems. Even the speedup is better in both cases, being the best of the studied algorithms for the MMDP with a value of 6.76$\times$. In the case of MMDP the speedup of the three algorithms was quite good although lineal speedup was not reached.

In summary, the speedup for the MMDP was quite good although lineal speedup was not achieved. In the case of SSP, Ethane G and B have not showed a very good speedup, and ssGA has showed even a worse speedup.  This fact could be explained by the huge difference among the power of the different hardware configurations used (remember that the reference point for speedup is the best performing processor).

\subsection{Evolution of the Fitness}

\begin{figure}[!t]
	\includegraphics[width=3.5in]{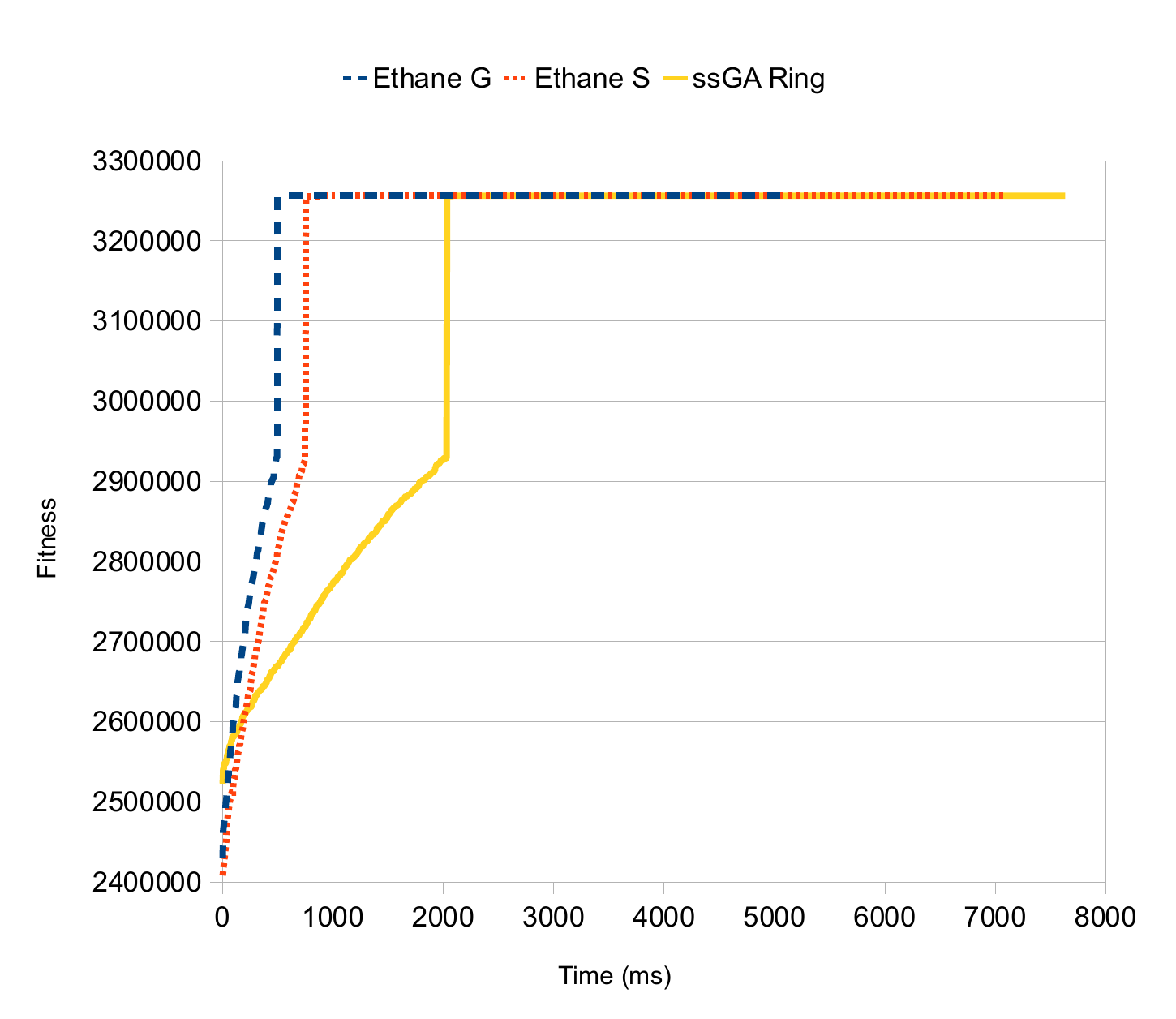}
	\caption{Evolution of fitness for SSP (time - ms - vs. fitness)}
	\label{subset_fitness}
\end{figure}

\begin{figure}[!t]
	\includegraphics[width=3.5in]{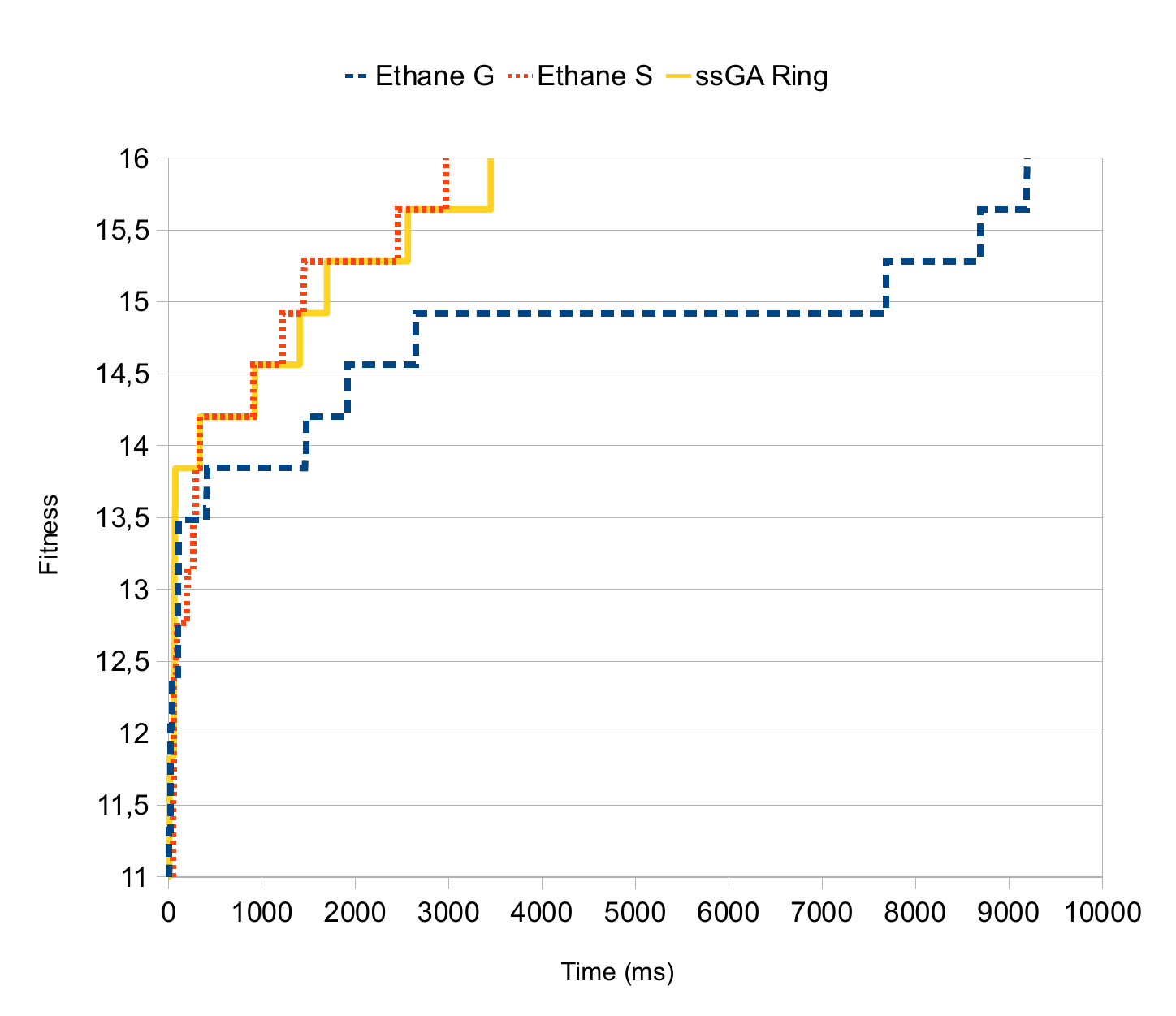}
	\caption{Evolution of fitness for SSP (time - ms - vs. fitness)}
	\label{mmdp_fitness}
\end{figure}


Figures \ref{subset_fitness} and \ref{mmdp_fitness} are showing the median execution of each algorithm for each problem. 

In the case of SSP, the Figure shows that both Ethane versions clearly outperforms the ssGA ring, converging quite faster. We can see how Ethane G performs even better than Ethane S for this problem. 

For the MMDP, Ethane S performed clearly better than Ethane G as we can see in Figure \ref{mmdp_fitness}. Ethane S outperformed the ssGA ring, but the difference is not as large as with the SSP.

\section{Conclusions and Future Work}\label{section7}

In this paper we have presented a new heterogeneous parallel search algorithm based on the structure of ethane. We have also shaped a general model for designing heterogeneous algorithms depending on the underlying heterogeneous platform, inspired in the structures of the hydrocarbons present in Nature.

We have performed a set of tests in order to assess the performance of our proposal, and compared it with a well-known state-of-the-art model, the ssGA unidirectional ring, and two well-known algorithms: SA and ssGA. Our tests have shown that our proposal can perform better in terms of time and numerical effort than the reference model, and Ethane is even able to find the solutions in a more robust/stable manner. Also the speedup of the proposed models is competitive with that of the reference model, obtaining quite good values even with the huge differences between the performance of the computers of the heterogeneous platform.

As future work we propose to extend and deeply analyze the HydroCM model, as well as to assess its performance with different configurations and real-life applications. Our goal is to offer a general model for gracefully matching computers of different powers to run different algorithms for efficiently solve the same problem, in a way that an heterogeneous platform does not constitute a problem but, on the contrary, could be used as a target platform for specialized new parallel algorithms.

\section*{Acknowledgment}

This work has been partially funded by the Spanish Ministry of Science and Innovation and FEDER under contract TIN2008-06491-C04-01 (the M* project). It has also been partially funded by the Andalusian Government under contract P07-TIC-03044 (DIRICOM project).

\balance


\begin{thebibliography}{1}

\bibitem{Alba0}
E.~Alba, ``Metaheuristics and Parallelism''. \hskip 1em plus
0.5em minus 0.4em\relax \emph{Parallel Metaheuristics: A new Class of Algorithms} Wiley-Interscience, pp. 79-103, 2005.
	
\bibitem{Alba}
E.~Alba, ``Parallel Heterogeneous Metaheuristics''. \hskip 1em plus
0.5em minus 0.4em\relax \emph{Parallel Metaheuristics: A new Class of Algorithms} Wiley-Interscience, pp. 395-422, 2005.

\bibitem{Alba3}
E.~Alba, F.~Luna, A.J.~Nebro, and J.M.~Troya, ``Parallel heterogeneous genetic algorithms for continuous optimization''. \hskip 1em plus
  0.5em minus 0.4em\relax Parallel Computing, Volume 30, Issues 5-6, pp 699-719, 2004.

\bibitem{Alba2}
E.~Alba, A.J.~Nebro, and J.~Troya, ``Heterogeneous Computing and Parallel Genetic Algorithms''.\hskip 1em plus
  0.5em minus 0.4em\relax \emph{Journal of Parallel and Distributed Computing}, 62: 1362-1385, 2002.

\bibitem{Alba4}
E.~Alba and J.M.~Troya, ``Analyzing synchronous and asynchronous parallel distributed genetic algorithms''. \hskip 1em plus
  0.5em minus 0.4em\relax \emph{Future Generation Computer Systems}, Volume 17, pp 451-465, 2001.
	
\bibitem{Branke}
J.~Branke, A.~Kamper, and H.~Schmeck, ``Distribution of Evolutionary Algorithms in Heterogeneous Networks''.\hskip 1em plus
  0.5em minus 0.4em\relax \emph{Lecture Notes in Computer Science}, Volume 3102/2004,pp. 923-934, 2004.

\bibitem{Goldberg}
D.E.~Goldberg, K.~Deb, and J.~Horn, ``Massively multimodality, deception and genetic algorithms''.\hskip 1em plus
  0.5em minus 0.4em\relax \emph{Parallel Problem Solving from Nature}, 2: 37-46, 1992.

\bibitem{Jelasity}
M.~Jelasity, ``A wave analysis of the subset sum problem''.\hskip 1em plus
  0.5em minus 0.4em\relax \emph{Proceedings of the Seventh International Conference on Genetic Algorithms}, San Francisco, CA, pp. 89-96, 1997.

\bibitem{Salto}
C.~Salto, and E.~Alba ``Designing Heterogeneous Distributed GAs by Efficient Self-Adapting the Migration Period''.\hskip 1em plus
  0.5em minus 0.4em\relax \emph{Applied Intelligence}, to appear, 2011.

\bibitem{Salto2}
C.~Salto, E.~Alba, and F.~Luna, ``Using Landscape Measures for the Online Tuning of Heterogeneous Distributed GAs''.\hskip 1em plus
  0.5em minus 0.4em\relax \emph{Genetic and Evolutionary Computation Conference}, to appear, 2011.

\bibitem{Syswerda}
G.~Syswerda, ``A study of reproduction in generational and steady-state genetic algorithms''.\hskip 1em plus
  0.5em minus 0.4em\relax \emph{Foundations of Genetic Algorithms}, Morgan Kauffman, pp. 94-101, 1991.

\bibitem{Yao}
X.~Yao, ``A new Simulated Annealing Algorithm''.\hskip 1em plus
  0.5em minus 0.4em\relax \emph{International Journal of Computer Mathematics}, 56: 161-168, 1995.

\end{thebibliography}
\end{document}